%% file: paper.tex
\documentclass[11pt,journal,compsoc,onecolumn]{IEEEtran}
\usepackage{cite}
\usepackage{times}
\usepackage{epsfig}
\usepackage[ruled,lined,linesnumbered,boxed]{algorithm2e}
\usepackage{color}
\usepackage{amsmath,graphicx,multirow}
\usepackage{bbm,amssymb}
\usepackage{dsfont}
\usepackage{algorithmic}

\def\t{{\bf \textrm{tr}}}

\begin{document}

\title{Image Annotation based on Deep Hierarchical Context Networks}
 
\author{Mingyuan Jiu, 
        Hichem Sahbi
 \thanks{M. Jiu is with School of Information Engineering, Zhengzhou University. Zhengzhou, 450001, China. Email: iemyjiu@zzu.edu.cn.} 
\thanks{H. Sahbi is with CNRS LIP6, UPMC Sorbonne University, Paris, 75005, France. Email: hichem.sahbi@sorbonne-universite.fr.}
}

\maketitle

\begin{abstract}
Context modeling is one of the most fertile subfields of visual recognition which aims at designing  discriminant image representations while incorporating their intrinsic and extrinsic relationships. However, the potential of context modeling is currently underexplored and most of the existing solutions are either context-free or restricted to simple handcrafted geometric relationships. \\ \noindent We introduce in this paper DHCN: a novel Deep Hierarchical Context Network that leverages different sources of contexts including geometric and semantic relationships. The proposed method is based on the minimization of an objective function mixing a fidelity term, a context criterion and a regularizer. The solution of this objective function defines the architecture of a bi-level hierarchical context network; the first level of this network captures scene geometry while the second one corresponds to semantic relationships. We solve this representation learning problem by training its underlying deep network whose parameters correspond to the most influencing bi-level contextual relationships and we evaluate its performances on image annotation using the challenging ImageCLEF benchmark. 
\end{abstract}


\IEEEpeerreviewmaketitle

\section{Introduction} \label{sec:intro}

\IEEEPARstart{I}{mage} annotation is one of the major challenges in computer vision which aims at assigning keywords (a.k.a labels or concepts) to images \cite{Bernard2003,sahbicassp11,Makadia2008,boujemaa2004visual,sahbiicip18,Goh2005,Qi2007,Guillaumin2009,sahbiclef08,Verma2012,sahbiicip09,Hecvpr2004,Zhang2018,Verma2013,sahbiiccv17,Jiu2015,sahbiphd,Jiu2016a,JiuPR2019,Vo2012,sahbiacmm2000,deng2014deep,Goodfellowetal2016,srivastava2015training,szegedy2015going,russakovsky2015imagenet,sahbiigarss12a,sahbisc2008,icassp2017b,sahbiarxiv2017,sahbiicassp2019}. The difficulty in image annotation stems from the extreme variability of the learned concepts and their versatile content which is usually described with handcrafted or learned representations~\cite{sahbiCI2005,sahbiicip2001,sahbifuzzy2005,Murthyicmr2015,WangJVCIR2017,Jiutip2017,ZhangIET2018,Cheng2018,Liu2018, ZhengPami2018, Bhagat2018,Belongie01shapematching,YLiupr2007,ZhangDPR2012,Wong2008,Kuroda2002,Cusano2004,NiuTIP2019,MaMTA2019,sahbiicassp13a,BelongieMalik2002,sahbijstars17,JinMTA2019,lowe1999object,girshick2014rich,Metzlercivr04,Lavrenko2003,Villegas2013,Duygulu2002}. However, due to its limited representational power, content is usually upgraded with context in order to capture both the intrinsic and the extrinsic properties of images\footnote{\scriptsize Intrinsic properties of images are usually related to scene structure or geometry while extrinsic properties refer to semantic relationships (such as ``image-to-image'' links in social networks).}. Indeed, while context-free models are effective when images (from the same concepts) are well clustered, they miserably fail when concepts exhibit a strong intra-class variability. In contrast, context-dependent solutions reduce the ratio between intra and inter class variability even when content of images --- belonging to the same concepts --- is corrupted \cite{Sahbi2011a,Grangier2008,ZhangBaiICCV2107,HungTsaiICCV2107,HungTsaiCVPR2107b,MartinsJVCIR2014,ArunIJMLC2017,ZhangEAAI2019}. \\
\begin{figure*}[t]
  \centering
 \scalebox{0.32}{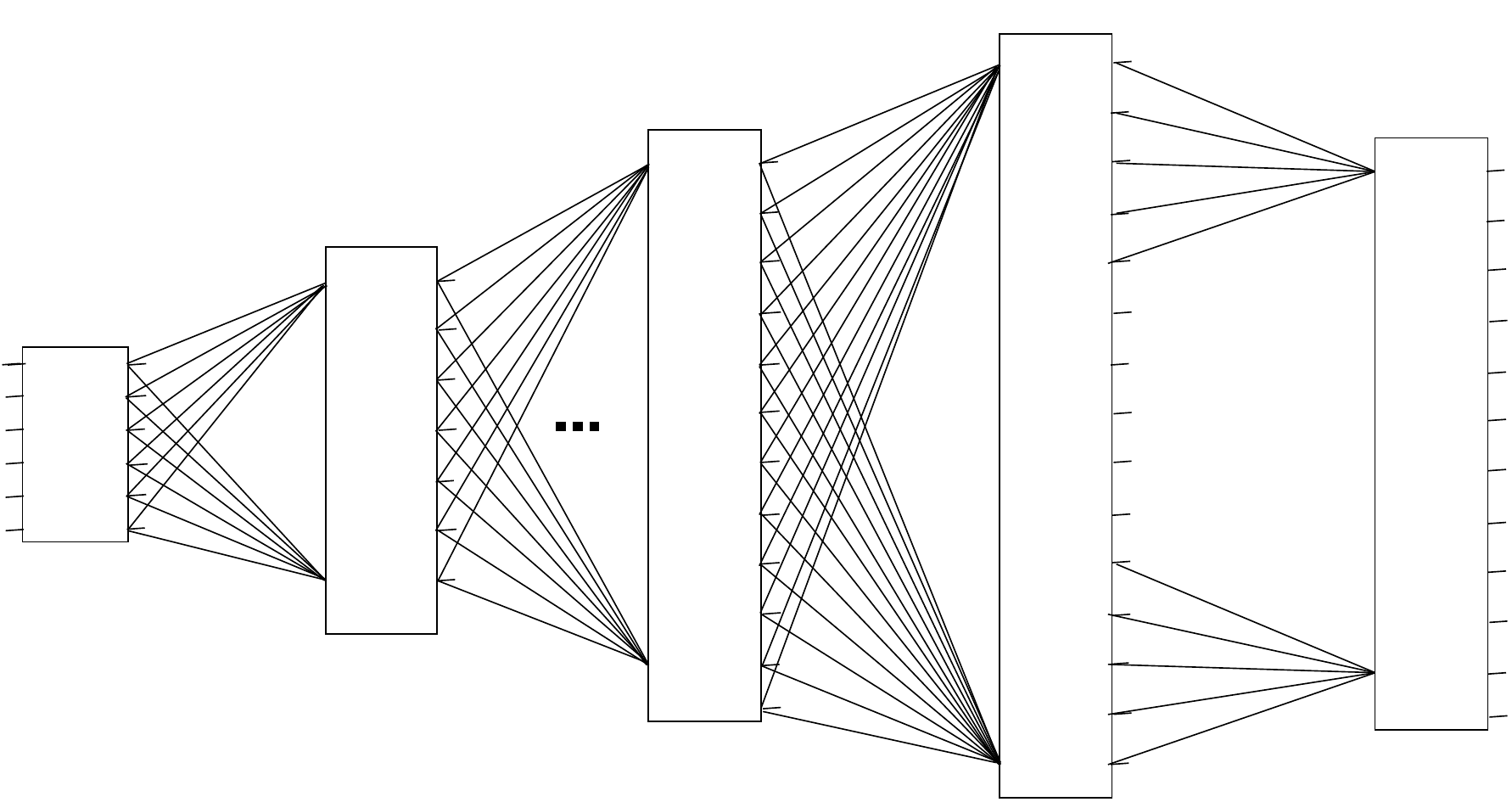} \hspace{0.5cm} \includegraphics[width=0.62\linewidth]{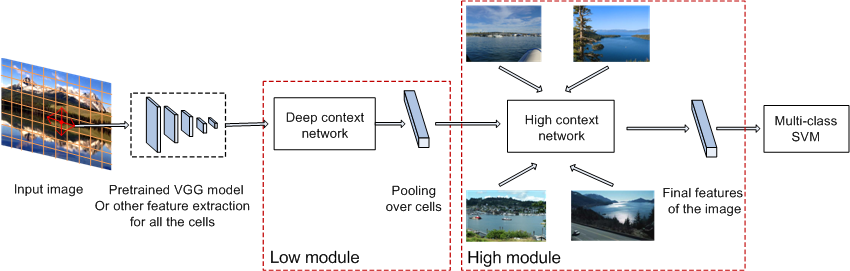}
  \caption{\scriptsize (Left) This figure shows the ``unfolded'' multi-layered context network with increasing dimensionality that captures larger and more influencing contexts. (Right) Hierarchical context learning framework including geometric and semantic contexts. In this diagram, we only show  four semantic neighbors for each image as an example {\bf (better to zoom the PDF version)}.  \label{fig:framework}}
\end{figure*}

Several existing methods leverage context prior to achieve image annotation and outperform context-free approaches by a significant margin. In these solutions, context is usually defined as a neighborhood system \cite{sahbiigarss2012a}, i.e., a set of geometric or statistical dependencies between low level primitives (such as interest points, regions, etc.) or semantic relationships. These relationships make it possible to model pairwise and high-order interactions between images and their primitives using well designed objective functions; several works follow this line including neighborhood embedding~\cite{Salakhutdinov2007JMLR} and spatially-constrained deep learning~\cite{Hadsell06,Jiuprl2014}. These methods learn functions that map neighboring data from the input (raw) space into a well designed feature space while maintaining their proximity. Other methods rely on structural regularization which integrates a priori knowledges into different penalization terms and constrain  the learned models to reflect these knowledges. Typical works include $\ell_1$-norm~\cite{Tibshirani1996}, $\ell_0$-norm, $\ell_{12}$-norm~\cite{Jacob2009ICML} and structural regularization~\cite{Bach2011cStatisticalScience} which usually define convex (globally optimal) problems. Variants  of these models consider prediction scores on labeled and unlabeled data for regularization (as in Laplacian SVMs~\cite{Belkin2006JMLR}) in order to diffuse labels from training to test data.  More recently, graph neural networks have attracted a particular attention as an extension of convolutional neural networks (CNNs)~\cite{LeCun98,Hinton2006,Krizhevsky2012,LeCun2015science,HeKaiMing2016cvpr}  to non-Euclidean domains~\cite{Defferrard2016NIPS, Kipf2017semi, OretegaIEEE2018, Wu2019arXiv, LeeCWcvpr2018, Knyazev2019bmvc} and have shown very promising performances on relational graph data. \\

\indent The success of all the aforementioned methods is very dependent on the relevance of the used neighborhood systems which are usually handcrafted and when learned they are restricted only to simple geometric relationships. In contrast, the solution proposed in this paper learns both  geometric and semantic contextual relationships in a unified framework. Our design principle relies on the context-dependent similarities introduced in~\cite{Sahbi2011,Sahbi2011a,Sahbi2013icvs,Jiuicpr2018,sahbicbmi08,Sahbi2015} but, in contrast to these works, considers learned bi-level contexts instead of handcrafted ones. Learning context translates into optimizing the adjacency matrices of the neighborhood system and this equivalently reduces to training a particular deep network whose parameters correspond to the most influencing (geometric and semantic) relationships in image annotation. With this approach, the representation of a given image is obtained recursively by aggregating (i) the representations of neighboring primitives (insides images) following the learned geometric context and (ii) those of neighboring images according to the learned semantic context. This results into a highly discriminant hierarchical representation as shown later in experiments.

\section{context-aware similarity networks} \label{sec:dcn}
Let ${\cal I}=\{{\cal I}_p\}_{p=1}^P$ denote  a collection of training images and ${\cal S}_p=\{ {\bf x}_1^p, \ldots, {\bf x}_n^p \}$ be a list of non-overlapping cells taken from a regular grid of ${\cal I}_p$; without a loss of generality, we assume $n$ constant for all images. A context-aware similarity (or kernel denoted as $\kappa$) is a symmetric and positive semi-definite (p.s.d) function that returns the resemblance between any two given cells ${\bf x}$, ${\bf x}'$ in ${\cal X}=\cup_p {\cal S}_p$. As designed subsequently, the particularity of $\kappa$ w.r.t many usual kernels (such as linear, RBF, etc. \cite{ShaweTaylor2004,barla2003histogram,Grauman2007,Lanckriet2004,lingsahbieccv2014,lingsahbiicip2014,sahbirr2002,sahbirr2004}) is that $\kappa({\bf x},{\bf x}')$ depends {\it not only} on the content of the cells $({\bf x},{\bf x}')$ {\it but also} on their context $\{{\cal N}_c({\bf x}) \times {\cal N}_c({\bf x}')\}_c$; here $\{{\cal N}_c({\bf x})\}_c$ corresponds to the neighborhood system, i.e., the set of neighbors of ${\bf x}$ with particular (learned) geometric relationships. The kernel $\kappa$ (or equivalently its gram matrix $\mathbf{K}$) is learned by minimizing the following objective function 
\begin{equation}
 \min_{\mathbf{K}} \t(-\mathbf{K}\mathbf{S}') - \alpha_1 \sum_{c=1}^C \t (\mathbf{K} \mathbf{P}_c \mathbf{K}' \mathbf{P}_c^{'} ) + \frac{\beta_1}{2} ||\mathbf{K}||^2_2,
\label{eq:kernelfunction}
\end{equation}
here $'$ and $\t$ denotes matrix transpose and the trace operator respectively, $\alpha_1 \geq 0$, $\beta_1 > 0$, $\kappa({\mathbf{x}, \mathbf{x}'})=\mathbf{K}_{\mathbf{x}, \mathbf{x}'}$ (with $\mathbf{K}_{\mathbf{x}, \mathbf{x}'}$ being an entry of $\mathbf{K}$) and $\mathbf{S}$ is a (context-free) visual similarity matrix between data in $\cal X$. In the above objective function the matrices $\{\mathbf{P}_c\}_c$ correspond to a neighborhood system $\{{\cal N}_c(.)\}_{c=1}^C$; each entry $\mathbf{P}_{c,\mathbf{x},\mathbf{x}'} \neq 0$ if $\mathbf{x}' \in {\cal N}_c({\bf x})$, otherwise $\mathbf{P}_{c,\mathbf{x},\mathbf{x}'}\leftarrow 0$. In practice, $C$ (with $C=4$) different types of neighbors are considered (top, bottom, left, right) and the initial spatial support of these neighbors $\{{\cal N}_c(\mathbf{x})\}_{c=1}^C$ corresponds to a disk with a radius $r$ around $\bf x$ (see more details about the setting of $r$ in experiments). Using $\kappa$, one may define the similarity between any two given images ${\cal I}_p$ and ${\cal I}_q$ using convolution which aggregates the similarities between all the pairs in ${\cal S}_p \times {\cal S}_q$ as ${\cal K}( {\cal S}_p, {\cal S}_q) = \sum_{i, j} \kappa ({\bf x}_i^p, {\bf x}_j^q)$. Note that ${\cal K}$ is also symmetric and p.s.d resulting from the closure of the positive semi-definiteness w.r.t the sum. \\

\noindent One may show that the solution of Eq.~\eqref{eq:kernelfunction} is recursively obtain as the fixed-point (denoted as $\tilde{\mathbf{K}}$) of 
\begin{equation}
\mathbf{K}^{(t+1)} = \mathbf{S} + \gamma_1 \sum_{c=1}^C \mathbf{P}_c \mathbf{K}^{(t)} \mathbf{P}_c^{'},
\label{eq:kernelsolution} 
\end{equation}
\noindent with $\gamma_1=\alpha_1/\beta_1$. Resulting from the p.s.d of $\{\mathbf{K}^{(t)}\}_t$ (thereby $\tilde{\mathbf{K}}$) and ${\cal K}$, the  maps associated to these kernels are respectively   
\begin{equation}
\begin{array}{lll}
  \mathbf{\Phi}^{(t+1)}& =& \Big( \mathbf{\Phi}^{'(0)} \  \   \gamma_1^{\frac{1}{2}} \mathbf{P}_1 \mathbf{\Phi}^{'(t)} \ \ldots \  \  \gamma_1^{\frac{1}{2}} \mathbf{P}_C \mathbf{\Phi}^{'(t)} \Big)' \\
\displaystyle \phi_{\cal K}({\cal S}_p) &=& \displaystyle \sum_{\mathbf{x} \in {\cal S}_p} \tilde{\mathbf{\Phi}}_{\bf x}, 
\end{array}
                             \label{eq:mapsolution}
\end{equation}
here $\tilde{\mathbf{\Phi}}_{\bf x}$ denotes the restriction of $\tilde{\mathbf{\Phi}}$ to ${\bf x}$ and $\mathbf{\Phi}^{(0)}$ is the map of the initial kernel $\mathbf{K}^{(0)}$; for instance, this initial map can be exactly set using the Kronecker tensor product for the polynomial kernel or approximated using KPCA for any other kernel (see more details in \cite{Sahbi2013icvs,Vedaldi2012,Maji2008}). Following the recursive form in Eq.~(\ref{eq:mapsolution}), it is easy to see that the latter is strictly equivalent to a multi-layered deep network (also referred to as {\it deep context network}) whose input is $\mathbf{\Phi}^{(0)}$, intermediate layers $\{\mathbf{\Phi}^{(t)}\}_t$, output $\displaystyle \phi_{\cal K}({\cal S}_p)$ and weights corresponding to the adjacency matrices $\{\mathbf{P}_c\}_c$ (see Fig.~\ref{fig:framework}, left); hence training this network makes it possible to learn the neighborhood system, i.e., the spatial (geometric) context.

\section{Deep Hierarchical context learning} \label{sec:hdcn}
 In this section, we extend the previous framework to build a deep hierarchical context network that learns not only geometric but also semantic relationships between images. This turns out to be more effective as shown later in experiments. 
 \subsection{Bi-level context learning} \label{sec:highlevel}
As describe earlier, context learning makes it possible to capture spatial relationships between image cells. While being already performant, this design focuses mainly on the geometric structure of images and ignores totally other types of relationships, namely semantic ones. The tenet in this extension is to consider an extra-level in context-aware similarity design that considers images similar not only when their learned representations $\{\phi_{\cal K}({\cal S}_p)\}_p$ are close but also when their semantic context is similar too. The notion of semantic context is inherently different but complementary w.r.t  the one used earlier; indeed, the semantic neighborhood system (now denoted as ${\cal N}_{\cal I}({\cal S}_p)$), associated to any given image ${\cal S}_p$, is defined as the set of images sharing semantic relations\footnote{For instance, one may consider these relations using similarity or links in social networks.} with ${\cal S}_p$.  Considering  $\mathbf{P}_{\cal I}$ as the adjacency matrix related to ${\cal N}_{\cal I}(.)$, and $\mathbf{K}_{\cal I}$ the targeted context-aware similarity (to learn), we find the latter by minimizing a variant of Eq.~(\ref{eq:kernelfunction})
\begin{equation} \label{eq:kernelimage}
 \min_{\mathbf{K}_{\cal I}} \t(-\mathbf{K}_{\cal I}\tilde{\mathbf{S}}') - \alpha_{2}  \t (\mathbf{K}_{\cal I} \mathbf{P}_{\cal I} \mathbf{K}_{\cal I}^{'} \mathbf{P}^{'}_{\cal I} ) + \frac{\beta_2}{2} ||\mathbf{K}_{\cal I}||^2_2,
\end{equation}
\noindent here $\alpha_2\geq 0$, $\beta_2>0$, $\mathbf{K}_{\cal I}$ is the learned similarity matrix for images in $\cal I$ and entries of $\tilde{\mathbf{S}}$ correspond to inner products of the obtained $\{\phi_{\cal K}({\cal S}_p)\}_p$ on the fixed-points of Eq.~(\ref{eq:mapsolution}). Similarly, one may show that the solution of Eq.~(\ref{eq:kernelimage}) can be recursively defined as $\mathbf{K}_{\cal I}^{(t+1)} = \tilde{\mathbf{S}} + \gamma_2  \mathbf{P}_{\cal I} \mathbf{K}_{\cal I}^{(t)}  \mathbf{P}_{\cal I}^{'}$ which is again a p.s.d kernel whose map is explicitly given by
\begin{equation}
\mathbf{\Phi}^{(t+1)}_{\cal I} = \Big( \mathbf{\Phi}^{'(0)}_{\cal I} \  \   \gamma_2^{\frac{1}{2}} \mathbf{P}_{\cal I} \mathbf{\Phi}^{'(t)}_{\cal I} \Big)',
\label{eq:imagemapsolution}
\end{equation}
\noindent with $\gamma_2=\alpha_2/\beta_2$. By combining the recursive forms in Eqs.~(\ref{eq:mapsolution}) and (\ref{eq:imagemapsolution}), one may define a deep context network (related to Eq.~\ref{eq:imagemapsolution}) on top of another one (related to Eq.~\ref{eq:mapsolution}); training the parameters  $\{\mathbf{P}_c\}_c$, $\mathbf{P}_{\cal I}$ of this complete deep hierarchical context network (DHCN) makes it possible to learn  bi-level contextual relationships where the first level captures low-order geometric relationships while the second level models high-order semantic links between images. The whole architecture is shown in Fig.~(\ref{fig:framework}, right).
\begin{table*}[t]
	\centering
	\resizebox{0.9\textwidth}{!}{	
	\begin{tabular}{|c|c|c|cc|cc|}
	\hline
	\multirow{2}{*}{Method} & \multirow{2}{*}{$r$} & \multirow{2}{*}{\textcolor{black}{$|\mathcal{N}_{\cal I}|$}} & \multicolumn{2}{c|}{BoW features} & \multicolumn{2}{c|}{VGG-CNN features} \\
    \cline{4-7}
	& & & Lin kernel map  & HI kernel map & Lin kernel map &  HI kernel map \\
	\hline
	CF (Context-free) & - & - & 39.7/24.4/46.6 & 41.3/25.1/49.5 & 45.3/30.8/56.4 & 45.5/30.1/57.9  \\
	\hline
	DFCN (Deep fixed context network~\cite{Sahbi2013icvs}) & 1 & - & 40.6/24.6/48.3 & 42.6/26.3/50.5 & 45.8/31.2/57.6 & 46.4/30.7/58.5  \\
	DLCN (Deep learned context network~\cite{Jiuicpr2018}) & 1 & - & 42.7/26.4/50.5 & 45.2/26.4/53.9 & 47.5/32.7/58.7 & 48.8/32.7/59.9 \\
	DHCN (proposed) & 1 & 10 & 54.6/43.2/64.8 & \textbf{55.5}/43.4/65.3 & 56.0/\textbf{44.8}/65.6 & 55.7/\textbf{44.7}/65.8 \\
	\hline
	DFCN (Deep fixed context network~\cite{Sahbi2013icvs}) & 5 & - & 41.0/25.3/48.9 & 42.9/26.7/51.3 & 46.8/31.8/57.9 & 46.9/31.1/58.7  \\
	DLCN (Deep learned context network~\cite{Jiuicpr2018}) & 5 & - & 44.0/26.6/52.0 & 45.6/26.2/54.0 & 47.9/33.2/58.8 & 48.4/32.7/59.5 \\
	DHCN (proposed) & 5 & 10 & 54.6/39.8/\textbf{64.9} & \textbf{55.5}/42.0/65.7 & 56.1/44.0/65.7 & \textbf{56.5}/43.8/\textbf{66.6} \\
	\cline{1-7}
	DHCN (proposed) & 1 & 15 & \textbf{54.7}/\textbf{43.6}/64.4 & 54.8/\textbf{43.6}/\textbf{66.0} & \textbf{56.2}/44.7/\textbf{66.3} & 56.0/44.4/66.1 \\	
	\hline
	\end{tabular} 
	}
	\caption{\scriptsize  The performance (in $\%$) of different methods in the test set of ImageCLEF. A triple $\cdot/\cdot/\cdot$ stands for MF-S/MF-C/mAP. In these experiments $r$ corresponds to the radius of the disk that supports geometric context while $|\mathcal{N}_{\cal I}|$ corresponds to the size of semantic context.\label{tab:imageclefresults}}
\end{table*} 

\begin{figure*}[thbp]
\begin{center}
\includegraphics[angle=0,width=1\linewidth]{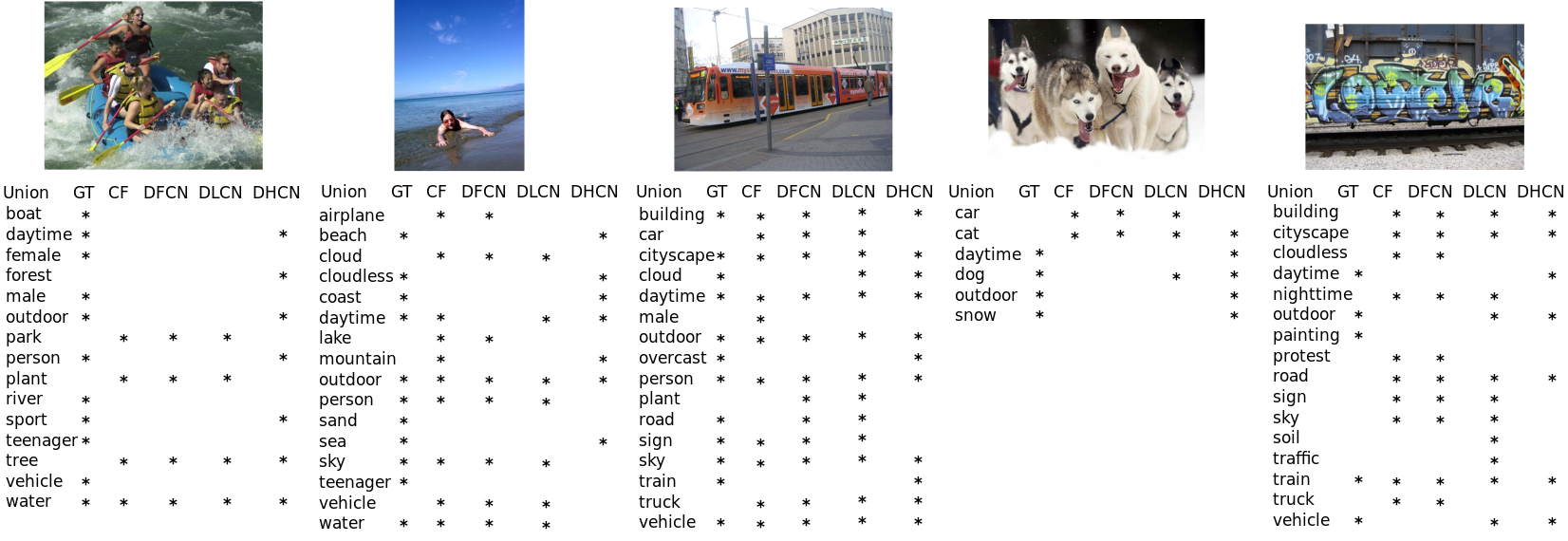}
\end{center}
\caption{\scriptsize Examples of annotation results using context-free representations (``CF''), deep context networks with fixed and learned contexts (resp. denoted ``DFCN'' and ``DLCN''), as well as deep hierarchical context network (``DHCN''). ``GT'' refers to ground-truth annotation while the stars mean the presence of a given concept in the test image.} \label{fig:annotationexamples}
\end{figure*}

\subsection{Optimization} \label{sec:hierlevel}
The two objective functions shown earlier define the complete architecture of the DHCN but training its parameters (and hence the context) requires another (supervised) loss. Considering a $K$-label classification task, a multi-class SVM layer (whose parameters denoted as $\{ w_k\}_k$) is stacked on top of DHCN for label prediction.  Let $\{({\cal I}_p,\mathbf{Y}_k^p)\}_p$ denote the training set of images and their labels with $\mathbf{Y}_k^p=+1$ iff ${\cal I}_p$ belongs to class $k$ and $\mathbf{Y}_k^p=-1$ otherwise. The supervised loss used to train our context matrices and SVM parameters is defined as 
\begin{equation}
 \min_{\{\mathbf{P}_c\}_c, \mathbf{P}_{\cal I}, w_k} \displaystyle   \sum_{k=1}^K \frac{1}{2} ||w_k||^2 + C_{k}  \sum_{p=1}^{P} \max(0, 1-\mathbf{Y}_k^p w_{k}' \mathbf{\Phi}_{\cal I}({\cal I}_p)). \label{eq:svmloss}
\end{equation}
\noindent We solve this problem using alternating optimization. First, we fix  $\{\mathbf{P}_c\}_c$ and $\mathbf{P}_{\cal I}$ and optimize the binary SVMs $\{f_k(.) = w_k' \mathbf{\Phi}_{\cal I}(.)\}_{k=1}^K$ using LIBSVM~\cite{Chang2011}. Then, we fix the learned SVMs and update the context parameters by gradient descent. Let $E$ denote the loss in Eq.~\eqref{eq:svmloss}, the gradient of $E$ w.r.t.~the final kernel map $\mathbf{\Phi}_{\cal I}({\cal I}_p)$ is given by
\begin{equation}
	\frac{\partial E}{\partial \mathbf{\Phi}_{\cal I}({\cal I}_p)} = - \sum_{p=1}^P \sum_{k=1}^K C_k \mathbf{Y}_k^p w_k \mathds{1}_{\{1-\mathbf{Y}_k^p w_{k}' \mathbf{\Phi}_{\cal I}({\cal I}_p) )\}}.
\label{equa:gradientl1}
\end{equation}
\noindent Using the chain rule~\cite{LeCun98}, we backpropagate this gradient to the previous layers in order to obtain all the gradients of $E$ w.r.t.~$\mathbf{P}_{\cal I}^{(t)}$ and $\{\mathbf{P}_c^{(t)}\}_c$ for $t= T-1, \ldots, 1$. Finally, we update the context matrices using gradient descent. These two iterative steps are repeated till convergence which is observed (in practice) in less than 100 iterations.
 
\section{Experiments} \label{sec:experiment}
\indent In this section, we apply the proposed DHCN to image annotation using the challenging ImageCLEF benchmark. The goal is to predict a list of keywords that best describes the visual content of images. This benchmark includes more than 250k images belonging to 95 concepts (also referred to as keywords); note that the latter are not exclusive, so one may assign multiple keywords to a given image when the scores of the underlying SVMs are positive. As the ground-truth has been released only on the dev set (of 1,000 images), we randomly split this set into two equally-sized subsets, one for training and another for evaluation. \\

\indent Each image in ImageCLEF is rescaled to a median dimension of $400\times500$  pixels and partitioned into a regular grid of $8\times10$ cells. Two types of features are used to describe the contents of the cells: i) Bag-of-Words (BoW) histogram with a SIFT code-book of 500 dimensions and ii) Deep VGG features pretrained on ImageNet (``imagenet-vgg-m-1024'')~\cite{Chatfield14}. This VGG-net is composed of five convolutional and three fully-connected layers and the output of the second fully-connected layer is used to describe the content of the cells in the regular grids. The performances are measured using the F-scores (harmonic means of recall and precision) both at  the concept and the sample levels (denoted respectively as MF-C and MF-S) as well as the mean average precision (mAP).

\begin{table}[htb]
	\centering
\resizebox{0.3\textwidth}{!}{	
	\begin{tabular}{c|ccc}
	\hline
	Kernel & MF-S & MF-C & mAP \\
	\hline
    GMKL~\cite{Varma2009} & 41.3 & 24.3 & 49.1 \\
    2LMKL~\cite{Zhuang2011a} & 45.0 & 25.8 & 54.0 \\
    LDMKL~\cite{Jiutip2017} & 47.8 & 30.0 & 58.6  \\
    DLCN~\cite{Jiuicpr2018}  & 48.8 & 32.7 & 59.9  \\ 
    DHCN (proposed) & \textbf{56.5} & \textbf{43.8} & \textbf{66.6}  \\ 
	\hline
	\end{tabular}
	}
    \caption{\scriptsize \textcolor{black}{Performance comparison w.r.t the most closely related work.} \label{tab:comparsionresults}}
\end{table}

\noindent In these experiments, we consider a six layer DHCN architecture corresponding to 2 (geometric context) + 1 (pooling) layers followed by 2 (semantic context)  + 1 (SVM) layers. Linear and histogram intersection (HI) maps are used as inputs to the DHCN and two settings of $r$ (the radius of the disk supporting the geometric context) are considered ($r=1$ and $r=5$). Note that the initial matrices $\{\mathbf{P}_c\}_c$ and $\mathbf{P}_{\mathcal{I}}$ (weights of DHCN) are normalized to be row-stochastic while  $\gamma_1$ and $\gamma_2$ are initially set to $1$. \\

\indent Table.~\ref{tab:imageclefresults} shows the performances of context-free networks (related to linear and HI kernel maps) vs. deep context networks with three settings: i) matrices in $\{\mathbf{P}_c\}_c$ are handcrafted (DFCN) ii) only $\{\mathbf{P}_c\}_c$ are learned (DLCN) and iii) both $\{\mathbf{P}_c\}_c$, $\mathbf{P}_{\mathcal{I}}$ are learned (DHCN). From all these results, we observe that the DHCN outperforms all the other settings by a large margin (for different features and kernel map initializations) compared to context-free and handcrafted deep context networks as well as learned ones (where only geometric context is learned);  globally, a more influencing impact on performances is observed with neighborhood systems learned with larger values of $r$ and $|{\cal N}_{\cal I}(.)|$. Finally, performance comparisons w.r.t the most related work are provided in Tab.~\ref{tab:comparsionresults} and some qualitative results in   Fig.~\ref{fig:annotationexamples}.

\section{Conclusion} \label{sec:conclu}

In this paper, we propose a deep hierarchical context network (DHCN) for image annotation. The method leverages two levels of contextual relationships; geometric and semantic. This is achieved by learning ``end-to-end'' the parameters of a deep context network whose architecture corresponds to the solution of an objective function that mixes  a content criterion that maximizes the similarity between visually close content, a context term  which restores the similarity when content is versatile and a regularizer that smooths the similarity and helps providing a closed-form solution. Training the parameters of this deep context network, using a supervised SVM loss, makes it possible to learn the most influencing geometric and semantic contextual relationships for image annotation. Experiments conducted on the challenging ImageCLEF benchmark, show that the proposed DHCN substantially enhances the performances of image annotation compared to shallow context-free as well as deep context networks with handcrafted or learned  (geometric only) contexts.  As a future work, we are currently investigating other priors on geometric and semantic relationships in order to further enhance the performances of image annotation. 

\section*{Acknowledgment}

 This work was supported by a grant from the National Natural Science Foundation of China (No.~61806180, U1804152), by a grant from Key Research Projects of Henan Higher Education Institutions in China (No.~19A520037), by a grant from Science and Technology Innovation Project of Zhengzhou (2019CXZX0037), and also in part by a grant from the research agency ANR (Agence Nationale de la Recherche) of France under the MLVIS project (ANR-11-BS02-0017).

\end{document}

%% file: deep.pdf_t
\begin{picture}(0,0)%
\includegraphics{deep.pdf}%
\end{picture}%
\setlength{\unitlength}{4144sp}%
\begingroup\makeatletter\ifx\SetFigFont\undefined%
\gdef\SetFigFont#1#2#3#4#5{%
  \reset@font\fontsize{#1}{#2pt}%
  \fontfamily{#3}\fontseries{#4}\fontshape{#5}%
  \selectfont}%
\fi\endgroup%
\begin{picture}(8135,4318)(915,-3230)
\put(1576,-646){\makebox(0,0)[lb]{\smash{{\SetFigFont{12}{14.4}{\rmdefault}{\mddefault}{\updefault}{\color[rgb]{0,0,0}$\phi_{0}$}%
}}}}
\put(3241,-151){\makebox(0,0)[lb]{\smash{{\SetFigFont{12}{14.4}{\rmdefault}{\mddefault}{\updefault}{\color[rgb]{0,0,0}$\phi_{1}$}%
}}}}
\put(4951,479){\makebox(0,0)[lb]{\smash{{\SetFigFont{12}{14.4}{\rmdefault}{\mddefault}{\updefault}{\color[rgb]{0,0,0}$\phi_{t-1}$}%
}}}}
\put(6931,929){\makebox(0,0)[lb]{\smash{{\SetFigFont{12}{14.4}{\rmdefault}{\mddefault}{\updefault}{\color[rgb]{0,0,0}$\phi_{t}$}%
}}}}
\put(8506,-1366){\makebox(0,0)[lb]{\smash{{\SetFigFont{12}{14.4}{\rmdefault}{\mddefault}{\updefault}{\color[rgb]{0,0,0}$\sum$}%
}}}}
\put(3691,-2536){\makebox(0,0)[lb]{\smash{{\SetFigFont{12}{14.4}{\rmdefault}{\mddefault}{\updefault}{\color[rgb]{0,0,0}${\bf P}_c^{(1)}$}%
}}}}
\put(5446,-3166){\makebox(0,0)[lb]{\smash{{\SetFigFont{12}{14.4}{\rmdefault}{\mddefault}{\updefault}{\color[rgb]{0,0,0}${\bf P}_c^{(t-1)}$}%
}}}}
\put(1891,-2176){\makebox(0,0)[lb]{\smash{{\SetFigFont{12}{14.4}{\rmdefault}{\mddefault}{\updefault}{\color[rgb]{0,0,0}${\bf P}_c^{(0)}$}%
}}}}
\end{picture}%

%% file: paper.bbl
\begin{thebibliography}{11}
 \bibitem{Murthyicmr2015}
V.~N. Murthy, S.~Maji, and R.~Manmatha,
\newblock ``Automatic image annotation using deep learning representations,''
\newblock in {\em International Conference on Multimedia Retrieval}, 2015, pp.
  603--606.

\bibitem{WangJVCIR2017}
R.~Wang, Y.~Xie, J.~Yang, L.~Xue, M.~Hu, and Q.~Zhang,
\newblock ``{Large scale automatic image annotation based on convolutional
  neural network},''
\newblock {\em Journal of Visual Communication and Image Representation}, vol.
  49, pp. 213--224, 2017.

\bibitem{Jiutip2017}
M.~Jiu and H.~Sahbi,
\newblock ``Nonlinear deep kernel learning for image annotation,''
\newblock {\em IEEE Transactions on Image Processing}, vol. 26(4), 2017.

\bibitem{ZhangIET2018}
J.~Zhang, Y.~Mu, S.~Feng, K.~Li, Y.~Yuan, and C.-H. Lee,
\newblock ``Image region annotation based on segmentation and semantic
  correlation analysis,''
\newblock {\em IET Image Processing}, vol. 12, no. 8, pp. 1331--1337, 2018.

\bibitem{Cheng2018}
Q.~Cheng, Q.~Zhang, P.~Fu, C.~Tu, and S.~Li,
\newblock ``{A survey and analysis on automatic image annotation},''
\newblock {\em Pattern Recognition}, vol. 79, pp. 242--259, 2018.

\bibitem{Liu2018}
Y.~Liu, K.~Wen, Q.~Gao, X.~Gao, and F.~Nie,
\newblock ``{SVM based multi-label learning with missing labels for image
  annotation},''
\newblock {\em Pattern Recognition}, vol. 78, pp. 307--317, 2018.
\bibitem{sahbicassp11}
  X. Li, H. Sahbi. Superpixel-based object class segmentation using conditional random fields. IEEE International Conference on Acoustics, Speech and Signal Processing (ICASSP). 2011.

\bibitem{ZhengPami2018}
L.~Zheng, Y.~Yang, and Q.~Tian,
\newblock ``{SIFT Meets CNN: A Decade Survey of Instance Retrieval},''
\newblock {\em IEEE Transactions on Pattern Analysis and Machine Intelligence},
  vol. 40, no. 5, pp. 1224--1244, 2018.

\bibitem{Bhagat2018}
P.~K. Bhagat and P.~Choudhary,
\newblock ``{Image annotation: Then and now},''
\newblock {\em Image and Vision Computing}, vol. 80, pp. 1--23, 2018.

\bibitem{Sahbi2011a}
H.~Sahbi and X.~Li,
\newblock ``Context-based support vector machines for interconnected image
  annotation,''
\newblock in {\em ACCV}, 2011, pp. 214--227.

\bibitem{Salakhutdinov2007JMLR}
R.~Salakhutdinov and G.~Hinton,
\newblock ``{Learning a nonlinear embedding by preserving class neighbourhood
  structure},''
\newblock {\em Journal of Machine Learning Research}, vol. 2, pp. 412--419.

\bibitem{Hadsell06}
R.~Hadsell, S.~Chopra, and Y.~LeCun,
\newblock ``{Dimensionality reduction by learning an invariant mapping},''
\newblock in {\em CVPR}, 2006, pp. 1735--1742.

\bibitem{Jiuprl2014}
M.~Jiu, C.~Wolf, G.~Taylor, and A.~Baskurt,
\newblock ``Human body part estimation from depth images via
  spatially-constrained deep learning,''
\newblock {\em Pattern Recognition Letteres}, vol. 50, pp. 122--129, 2014.

\bibitem{Tibshirani1996}
R.~Tibshirani,
\newblock ``{Regression shrinkage and selection via the lasso},''
\newblock {\em Journal of the Royal Statistical Society. Series B}, vol. 58,
  no. 1, pp. 267--288, 1994.

\bibitem{Jacob2009ICML}
L.~Jacob, G.~Obozinski, and J.~P. Vert,
\newblock ``{Group lasso with overlap and graph lasso},''
\newblock {\em ICML}, 2009.

\bibitem{sahbicbmi08}
H. Sahbi, JY. Audibert, J. Rabarisoa, R. Keriven. Object recognition and retrieval by context dependent similarity kernels. International Workshop on Content-Based Multimedia Indexing, 216-223, 2008.  


\bibitem{Bach2011cStatisticalScience}
F.~Bach, R.~Jenatton, J.~Mairal, and G.~Obozinski,
\newblock ``{Structured sparsity through convex optimization},''
\newblock {\em Statistical Science}, vol. 27, no. 4, pp. 1--27, 2011.

\bibitem{Belkin2006JMLR}
M.~Belkin, P.~Niyogi, and V.~Sindhwani,
\newblock ``{A geometric framework for learning from labeled and unlabeled
  examples},''
\newblock {\em Journal of Machine Learning Research}, vol. 7, pp. 2399--2434,
  2006.


\bibitem{Sahbi2015}
H.~Sahbi, ``Imageclef annotation with explicit context-aware kernel maps,''
  \emph{International Journal of Multimedia Information Retrieval}, pp.
  113--128, 2015.



\bibitem{Sahbi2011}
H.~Sahbi, J.-Y. Audibert, and R.~Keriven,
\newblock ``Context-dependent kernels for object classification,''
\newblock {\em IEEE Transactions on Pattern Analysis and Machine Intelligence},
  vol. 33, pp. 699--708, 2011.


\bibitem{LeCun98}
Y.~LeCun, L.~Botto, Y.~Bengio, and P.~Haffner,
\newblock ``Gradient-based learning applied to document recognition,''
\newblock {\em Proceedings of IEEE}, vol. 86, no. 11, pp. 2278--2324, 1998.

\bibitem{Hinton2006}
G.~Hinton, S.~Osindero, and Y-W. Teh,
\newblock ``A fast learning algorithm for deep belief nets,''
\newblock {\em Neural Computation}, vol. 18(7), pp. 1527--1554, 2006.

\bibitem{Krizhevsky2012}
A.~Krizhevsky, I.~Sutskever, and G.~E. Hinton,
\newblock ``Imagenet classification with deep convolutional neural networks,''
\newblock in {\em NIPS}, 2012.


\bibitem{Sahbi2013icvs}
H.~Sahbi,
\newblock ``Explicit context-aware kernel map learning for image annotation,''
\newblock in {\em ICVS}, 2013.

\bibitem{LeCun2015science}
Y.~LeCun, Y.~Bengio, and G.~Hinton,
\newblock ``{Deep learning},''
\newblock {\em Nature}, vol. 521, pp. 436--444, 2015.

\bibitem{HeKaiMing2016cvpr}
K.~He, X.~Zhang, S.~Ren, and J.~Sun,
\newblock ``Deep residual learning for image recognition,''
\newblock in {\em 2016 IEEE Conference on Computer Vision and Pattern
  Recognition (CVPR)}, Jun 2016, pp. 770--778.

\bibitem{Defferrard2016NIPS}
M.~Defferrard, X.~Bresson, and P.~Vandergheynst,
\newblock ``{Convolutional Neural Networks on Graphs with Fast Localized
  Spectral Filtering},''
\newblock in {\em NIPS}, 2016.

\bibitem{Kipf2017semi}
T.~N. Kipf and M.~Welling,
\newblock ``Semi-supervised classification with graph convolutional networks,''
\newblock in {\em International Conference on Learning Representations (ICLR)},
  2017.
\bibitem{Jiuicpr2018}
M.~Jiu, H.~Sahbi, and L.~Qi,
\newblock ``{Deep Context Networks for Image Annotation},''
\newblock in {\em Proceedings - International Conference on Pattern
  Recognition}, 2018, pp. 2422--2427.


\bibitem{OretegaIEEE2018}
A.~Ortega, P.~Frossard, J.~Kovacevic, J.~M. Moura, and P.~Vandergheynst,
\newblock ``{Graph Signal Processing: Overview, Challenges, and
  Applications},''
\newblock {\em Proceedings of the IEEE}, vol. 106, no. 5, pp. 808--828, 2018.

\bibitem{boujemaa2004visual}
N.~Boujemaa, F.~Fleuret, V.~Gouet, and H.~Sahbi, ``Visual content extraction
  for automatic semantic annotation of video news,'' in \emph{the proceedings
    of the SPIE Conference, San Jose, CA}, vol.~6, 2004.
  

\bibitem{Wu2019arXiv}
Z.~Wu, S.~Pan, F.~Chen, Long G., C.~Zhang, and P.~S. Yu,
\newblock ``A comprehensive survey on graph neural networks,''
\newblock {\em CoRR}, vol. abs/1901.00596, 2019.

\bibitem{sahbiicip18}
  H. Sahbi. Scene Decoding with Finite State Machines. 25th IEEE International Conference on Image Processing (ICIP), 485-489, 2018.

\bibitem{LeeCWcvpr2018}
C.-W. Lee, W.~Fang, C.-K. Yeh, and Y.-C. Wang,
\newblock ``{Multi-Label Zero-Shot Learning with Structured Knowledge
  Graphs},''
\newblock in {\em CVPR}, 2018, pp. 1576--1585.

\bibitem{Knyazev2019bmvc}
B.~Knyazev, X.~Lin, M.~Amer, and G.~Taylor,
\newblock ,''
\newblock in {\em BMVC}, 2019.


\bibitem{Chang2011}
C.-C. Chang and C.-J. Lin,
\newblock ``Libsvm: A library for support vector machines,''
\newblock {\em ACM Transactions on Intelligent Systems and Technology}, vol. 2,
  pp. 1--27, 2011.
\bibitem{sahbiclef08}
S. Tollari, P. Mulhem, M. Ferecatu, H. Glotin, M. Detyniecki, P. Gallinari, H. Sahbi,  Z-Q. Zhao. A comparative study of diversity methods for hybrid text and image retrieval approaches. In Workshop of the Cross-Language Evaluation Forum for European Languages, pp. 585-592. Springer, Berlin, Heidelberg, 2008.


\bibitem{Chatfield14}
K.~Chatfield, K.~Simonyan, A.~Vedaldi, and A.~Zisserman,
\newblock ``Return of the devil in the details: Delving deep into convolutional
  nets,''
\newblock in {\em BMVC}, 2014.

\bibitem{sahbisc2008}
H. Sahbi. "A particular Gaussian mixture model for clustering and its application to image retrieval." Soft Computing 12.7 (2008): 667-676.




\bibitem{icassp2017b}
M. Jiu and H. Sahbi. Deep kernel map networks for image annotation. IEEE International Conference on Acoustics, Speech and Signal Processing (ICASSP), 2016.


\bibitem{Varma2009}
M.~Varma and B.~Babu,
\newblock ``More generality in efficient multiple kernel learning,''
\newblock in {\em ICML}, 2009.

\bibitem{Zhuang2011a}
J.~Zhuang, I.~Tsang, and S.~Hoi,
\newblock ``Two-layer multiple kernel learning,''
\newblock in {\em ICML}, 2011, pp. 909--917.
\bibitem{sahbiicip09}
  M. Ferecatu, H. Sahbi. Multi-view object matching and tracking using canonical correlation analysis. 16th IEEE International Conference on Image Processing (ICIP), 2109-2112, 2009.


\bibitem{Bernard2003}
K.~Barnard, P.~Duygulu, N.~de~Freitas, and D.~Forsyth, ``Matching words and
pictures,'' \emph{JMLR}, vol.~3, 2003.


  
\bibitem{Grangier2008}
D.~Grangier and S.~Bengio, ``A discriminative kernel-based approach to rank  images from text queries,'' \emph{IEEE Transactions on Pattern Analysis and
  Machine Intelligence}, vol.~30, 2008.

  
\bibitem{ZhangBaiICCV2107}
Y.~Zhang, M.~Bai, P.~Kohli, S.~Izadi, and J.~Xiao, ``Deepcontext:  Context-encoding neural pathways for 3d holistic scene understanding,'' in
  \emph{ICCV}, 2017.

\bibitem{sahbiiccv17}
H. Sahbi. Coarse-to-fine deep kernel networks. Proceedings of the IEEE International Conference on Computer Vision, 1131-1139, 2017.



\bibitem{HungTsaiICCV2107}
W.-C. Hung, Y.-H. Tsai, X.~Shen, Z.~Lin, K.~Sunkavalli, X.~Lu, and M.-H. Yang,
  ``Scene parsing with global context embedding,'' in \emph{ICCV}, 2017.

\bibitem{HungTsaiCVPR2107b}
D.~Li, X.~Chen, Z.~Zhang, and K.~Huang, ``Learning deep context-aware features  over body and latent parts for person re-identification,'' in \emph{CVPR},
2017.

\bibitem{MartinsJVCIR2014}
P.~Martins, P.~Carvalho, and C.~Gatta, ``{Context-aware features and robust
  image representations},'' \emph{Journal of Visual Communication and Image
  Representation}, vol.~25, no.~2, pp. 339--348, 2014.

\bibitem{Jiu2015}
M.~Jiu and H.~Sahbi, ``Semi supervised deep kernel design for image   annotation,'' in \emph{ICASSP}, 2015.


\bibitem{ArunIJMLC2017}
K.~S. Arun and V.~K. Govindan, ``{A context-aware semantic modeling framework
  for efficient image retrieval},'' \emph{International Journal of Machine
  Learning and Cybernetics}, vol.~8, no.~4, pp. 1259--1285, 2017.



\bibitem{ZhangEAAI2019}
J.~Zhang, T.~Tao, Y.~Mu, H.~Sun, D.~Li, and Z.~Wang, ``{Web image annotation
  based on Tri-relational Graph and semantic context analysis},''
  \emph{Engineering Applications of Artificial Intelligence}, vol.~81, no. June
  2018, pp. 313--322, 2019.



\bibitem{Hecvpr2004}
X.~He, R.~Zimel, and M.~Carreira, ``Multiscale conditional random fields for   image labeling,'' in \emph{CVPR}, 2004.

  \bibitem{sahbiphd}
H. Sahbi.  Coarse-to-fine support vector machines for hierarchical face detection. PhD thesis, Versailles University, 2003. 


\bibitem{Zhang2018}
W.~Zhang, H.~Hu, and H.~Hu, ``{Neural ranking for automatic image  annotation},'' \emph{Multimedia Tools and Applications}, vol.~77, no.~17, pp.
  22\,385--22\,406, 2018.

\bibitem{Verma2013}
Y.~Verma and C.~Jawahar, ``Exploring svm for image annotation in presence of  confusing labels,'' in \emph{BMVC}, 2013.

  
\bibitem{Makadia2008}
A.~Makadia, V.~Pavlovic, and S.~Kumar, ``A new baseline for image annotation,''
in \emph{ECCV}, 2008, pp. 316--329.

  
\bibitem{Goh2005}
K.~Goh, E.~Chang, and B.~Li, ``Using one-class and two-class svms for
  multiclass image annotation,'' \emph{IEEE transactions on Knowledge and Data
  Engineering}, vol.~17, 2005.

\bibitem{Qi2007}
X.~Qi and Y.~Han, ``Incorporating multiple svms for automatic image
  annotation,'' \emph{IEEE Transactions on Knowledge and Data Engineering},
  vol.~40, 2007.

\bibitem{Jiu2016a}
M.~Jiu and H.~Sahbi, ``Laplacian deep kernel learning for image annotation,'' in  \emph{ICASSP}, 2016.


\bibitem{Guillaumin2009}
M.~Guillaumin, T.~Mensink, J.~Verbeek, and C.~Schmid, ``Tagprop: Discriminative
  metric learning in nearest neighbor models for image auto-annotation,'' in
  \emph{ICCV}, 2009, pp. 316--329.
  

\bibitem{Verma2012}
Y.~Verma and C.~Jawahar, ``Image annotation using metric learning in semantic
neighbourhoods,'' in \emph{ECCV}, 2012.





\bibitem{deng2014deep}
L.~Deng, D.~Yu, et~al.
\newblock Deep learning: methods and applications.
\newblock {\em Foundations and Trends{\textregistered} in Signal Processing},
  7(3--4):197--387, 2014.

 \bibitem{Goodfellowetal2016}
 I.~Goodfellow, Y.~Bengio, and A.~Courville.
 \newblock {\em Deep Learning}.
 \newblock MIT Press, 2016.
 
\bibitem{JiuPR2019}
M.~Jiu and H.~Sahbi, ``Deep representation design from deep kernel networks,''  \emph{Pattern Recognition}, vol.~88, pp. 447--457, 2019.





 \bibitem{srivastava2015training}
 R.~K. Srivastava, K.~Greff, and J.~Schmidhuber.
 \newblock Training very deep networks.
 \newblock In {\em Advances in neural information processing systems}, pages
 2377--2385, 2015.
 
   \bibitem{szegedy2015going}
 C.~Szegedy, W.~Liu, Y.~Jia, P.~Sermanet, S.~Reed, D.~Anguelov, D.~Erhan,
   V.~Vanhoucke, and A.~Rabinovich.
 \newblock Going deeper with convolutions.
 \newblock In {\em Proceedings of the IEEE conference on computer vision and
   pattern recognition}, pages 1--9, 2015.

\bibitem{Vo2012}
P.~Vo and H.~Sahbi, ``Transductive kernel map learning and its application to  image annotation,'' in \emph{BMVC}, 2012.


 \bibitem{russakovsky2015imagenet}
 O.~Russakovsky, J.~Deng, H.~Su, J.~Krause, S.~Satheesh, S.~Ma, Z.~Huang,
   A.~Karpathy, A.~Khosla, M.~Bernstein, et~al.
 \newblock Imagenet large scale visual recognition challenge.
 \newblock {\em International Journal of Computer Vision}, 115(3):211--252,
 2015.


\bibitem{sahbiacmm2000}
H. Sahbi and N. Boujemaa. "From coarse to fine skin and face detection." Proceedings of the eighth ACM international conference on Multimedia. 2000.


\bibitem{sahbiarxiv2017}
A. Dutta and H. Sahbi. "High order stochastic graphlet embedding for graph-based pattern recognition." arXiv preprint arXiv:1702.00156 (2017).

\bibitem{sahbiicassp2019}
A. Mazari and H. Sahbi. "Deep Temporal Pyramid Design for Action Recognition." ICASSP 2019-2019 IEEE International Conference on Acoustics, Speech and Signal Processing (ICASSP). IEEE, 2019.

  
\bibitem{Belongie01shapematching}
S.~Belongie, J.~Malik, and J.~Puzicha, ``Shape matching and object recognition
  using shape contexts,'' \emph{IEEE Transactions on Pattern Analysis and
  Machine Intelligence}, vol.~24, pp. 509--522, 2001.




\bibitem{sahbifuzzy2005}
H. Sahbi and N. Boujemaa. "Validity of fuzzy clustering using entropy regularization." The 14th IEEE International Conference on Fuzzy Systems, 2005. FUZZ'05.. IEEE, 2005.



\bibitem{YLiupr2007}
Y.~Liu, D.~Zhang, and G.~Lu, ``A survey of content-based image retrieval with
  high-level semantics,'' \emph{Pattern Recognition}, vol.~40, 2007.


\bibitem{ZhangDPR2012}
D.~Zhang, M.~Islam, and G.~Lu, ``A review on automatic image annotation
  techniques,'' \emph{Pattern Recognition}, vol.~45, 2012.

\bibitem{sahbiicip2001}
H. Sahbi and N. Boujemaa. "Robust matching by dynamic space warping for accurate face recognition." Proceedings 2001 International Conference on Image Processing (Cat. No. 01CH37205). Vol. 1. IEEE, 2001.

  


\bibitem{Wong2008}
R.~Wong and C.~Leung, ``Automatic semantic annotation of real-world web
  images,'' \emph{IEEE Transactions on Pattern Analysis and Machine
  Intelligence}, vol.~30, 2008.


\bibitem{Kuroda2002}
K.~Kuroda and M.~Hagiwara, ``An image retrieval system by impression words and
  specific object names?iris,'' \emph{Neurocomputing}, vol.~43, 2002.

  
\bibitem{Cusano2004}
C.~Cusano, G.~Ciocca, and S.~R., ``Image annotation using svm,'' in
  \emph{Proceedings of the Internet Image IV, vol. 5304, SIPE}, 2004.

\bibitem{sahbiCI2005}
H. Sahbi and N. Boujemaa. "Fuzzy clustering: Consistency of entropy regularization." Computational Intelligence, Theory and Applications. Springer, Berlin, Heidelberg, 2005. 95-107.


\bibitem{NiuTIP2019}
Y.~Niu, Z.~Lu, J.-R. Wen, T.~Xiang, and S.-F. Chang, ``{Multi-modal multi-scale
  deep learning for large-scale image annotation},'' \emph{IEEE Transactions on
  Image Processing}, vol.~28, no.~4, pp. 1720--1731, 2019.

\bibitem{sahbiigarss2012a}
N. Bourdis, D. Marraud, and H. Sahbi. "Spatio-temporal interaction for aerial video change detection." 2012 IEEE International Geoscience and Remote Sensing Symposium. IEEE, 2012.

\bibitem{sahbiigarss12a}
N. Bourdis, D. Marraud, H. Sahbi.  "Camera pose estimation using visual servoing for aerial video change detection." 2012 IEEE International Geoscience and Remote Sensing Symposium. IEEE, 2012.


\bibitem{MaMTA2019}
Y.~Ma, Y.~Liu, Q.~Xie, and L.~Li, ``{CNN-feature based automatic image
  annotation method},'' \emph{Multimedia Tools and Applications}, vol.~78,
  no.~3, pp. 3767--3780, 2019.



\bibitem{sahbiicassp13a}
E. Benhaim, H. Sahbi, and G.  Vitte. "Designing relevant features for visual speech recognition." 2013 IEEE International Conference on Acoustics, Speech and Signal Processing. IEEE, 2013. 


\bibitem{BelongieMalik2002}
S.~Belongie, J.~Malik, and J.~Puzicha, ``{Shape Matching and Object Recognition
  Using Shape Contexts},'' \emph{IEEE Transactions on Pattern Analysis and
  Machine Intelligence}, vol.~24, no.~24, pp. 509--521, 2002.

\bibitem{sahbijstars17}
Q. Oliveau, H. Sahbi. Learning attribute representations for remote sensing ship category classification.  IEEE Journal of Selected Topics in Applied Earth Observations and Remote Sensing, 2017. 

\bibitem{JinMTA2019}
C.~Jin, Q.~M. Sun, and S.~W. Jin, ``{A hybrid automatic image annotation
  approach},'' \emph{Multimedia Tools and Applications}, vol.~78, no.~9, pp.
  11\,815--11\,834, 2019.

\bibitem{lowe1999object}
D.~G. Lowe \emph{et~al.}, ``Object recognition from local scale-invariant   features.'' in \emph{iccv}, vol.~99, no.~2, 1999, pp. 1150--1157.

\bibitem{lingsahbieccv2014}
L. Wang, H. Sahbi. Nonlinear Cross-View Sample Enrichment for Action Recognition. European Conference on Computer Vision. Springer, 2014.

 \bibitem{girshick2014rich}
 R.~Girshick, J.~Donahue, T.~Darrell, and J.~Malik.
 \newblock Rich feature hierarchies for accurate object detection and semantic
   segmentation.
 \newblock In {\em Proceedings of the IEEE CVPR}, pages 580--587, 2014.


\bibitem{Metzlercivr04}
D.~Metzler and R.~Manmatha, ``A inference network approach to image
  retrieval,'' in \emph{International Conference on Image and Video Retrieval
    (CIVR)}, 2004, pp. 42--50.

\bibitem{Lavrenko2003}
V.~Lavrenko, R.~Manmatha, and J.~Jeon, ``A model for learning the semantics of
pictures,'' in \emph{NIPS}, 2003.

\bibitem{Villegas2013}
M.~Villegas, R.~Paredes, and T.~B., ``Overview of the imageclef 2013 scalable
  concept image annotation subtask,'' in \emph{CLEF}, 2013.


\bibitem{Duygulu2002}
P.~Duygulu, K.~Barnard, N.~de~Freitas, and D.~Forsyth, ``Object recognition as
  machine translation: Learning a lexicon for a fixed image vocabulary,'' in
  \emph{ECCV}, 2002.





\bibitem{lingsahbiicip2014}
L. Wang, H. Sahbi. Bags-of-Daglets for Action Recognition. IEEE International Conference on Image Processing (ICIP), 2014.



 
\bibitem{Vedaldi2012}
A.~Vedaldi and A.~Zisserman, ``Efficient additive kernels via explicit feature  maps,'' \emph{IEEE Transactions on Pattern Analysis and Machine
  Intelligence}, vol.~34, 2012.

\bibitem{Maji2008}
S.~Maji, A.~Berg, and J.~Malik, ``Classification using intersection kernel support vector machines is efficient,'' in \emph{CVPR}, 2008.


\bibitem{ShaweTaylor2004}
J.~Shawe-Taylor and N.~Cristianini, ``Kernel methods for pattern analysis,''
  \emph{Cambriage University Press}, 2004.


\bibitem{barla2003histogram}
A.~Barla, F.~Odone, and A.~Verri, ``Histogram intersection kernel for image classification,'' in \emph{Proceedings 2003 international conference on image
  processing (Cat. No. 03CH37429)}, vol.~3.\hskip 1em plus 0.5em minus
0.4em\relax IEEE, 2003, pp. III--513.



\bibitem{Grauman2007}
K.~Grauman and T.~Darrell, ``The pyramid match kernel: Efficient learning with   sets of features,'' \emph{JMLR}, vol.~8, pp. 725--760, 2007.

\bibitem{sahbirr2002}
H. Sahbi and F. Fleuret. "Scale-invariance of support vector machines based on the triangular kernel." (2002).

  
\bibitem{Lanckriet2004}
G.~Lanckriet, N.~Cristianini, P.~Bartlett, L.~E. Ghaoui, and M.~I. Jordan,   ``Learning the kernel matrix with semi-definite programming,'' \emph{JRML},
  vol.~5, pp. 27--72, 2004.

\bibitem{sahbirr2004}
H. Sahbi and F. Fleuret. "Kernel methods and scale invariance using the triangular kernel." (2004).






\end{thebibliography}
